\documentclass{article}

\usepackage{arxiv}

\usepackage[utf8]{inputenc} 
\usepackage[T1]{fontenc}    
\usepackage{hyperref}       
\usepackage{url}            
\usepackage{booktabs}       
\usepackage{amsfonts}       
\usepackage{nicefrac}       
\usepackage{microtype}      
\usepackage{graphicx}
\usepackage{natbib}
\usepackage{doi}
\usepackage{listings}
\usepackage{amsmath}
\usepackage{multirow}

\title{BibRank: Automatic Keyphrase Extraction Platform Using~Metadata}

\author{ Abdelrhman Eldallal \\
	Institute of Computer Science\\
	 University of Tartu\\
	Narva mnt 18, 51009 Tartu, Estonia \\
	\texttt{abdelrhman.d@aucegypt.edu} \\
	\And
	\href{https://orcid.org/0000-0002-3664-5367}{\includegraphics[scale=0.06]{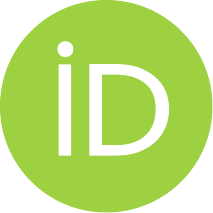}\hspace{1mm}Eduard Barbu} \\
	Institute of Computer Science\\
	 University of Tartu\\
	Narva mnt 18, 51009 Tartu, Estonia \\
	\texttt{eduard.barbul@ut.ee} \\
}
\renewcommand{\shorttitle}{\textit{BibRank} }

\begin{document}
\date{} 
\maketitle

\begin{abstract}
	Automatic Keyphrase Extraction involves identifying essential phrases in a document. These keyphrases are crucial in various tasks such as document classification, clustering, recommendation, indexing, searching, summarization, and text simplification. This paper introduces a platform that integrates keyphrase datasets and facilitates the evaluation of keyphrase extraction algorithms. The platform includes BibRank, an automatic keyphrase extraction algorithm that leverages a rich dataset obtained by parsing bibliographic data in BibTeX format. BibRank combines innovative weighting techniques with positional, statistical, and word co-occurrence information to extract keyphrases from documents. The platform proves valuable for researchers and developers seeking to enhance their keyphrase extraction algorithms and advance the field of natural language processing.
\end{abstract}


\section{Introduction}

The internet hosts an extensive collection of scientific documents, numbering in the tens of millions. Google Scholar, a web-based search engine dedicated to academic research, strives to provide comprehensive access to scholarly literature across various disciplines. A study \cite{10.1007/s11192-018-2958-5} reported that by the end of 2018, Google Scholar had indexed approximately 400 million articles. Keyphrases considered concise summaries of documents, aid information retrieval, indexing, and collection browsing. Automatic keyphrase extraction is the process of automatically identifying essential phrases within a document. Keyphrases find application in document clustering, classification, summarization, recommendation systems, and question answering. Automatic keyphrase extraction methods have been developed in domains such as social media, medicine, law, and agriculture, where they support specialized systems for organizing and retrieving information \cite{7805062} \cite{AlamiMerrouni2019AutomaticKE}.

Automatic keyphrase extraction methods can be categorized into unsupervised, supervised, and semi-supervised. Unsupervised techniques, which are domain-dependent, do not require labeled training data. On the other hand, supervised methods rely on manually annotated data, while semi-supervised ones strike a balance by requiring less annotated data compared to supervised methods.

This paper introduces a downloadable platform that integrates keyphrase datasets in BibTeX format and facilitates the evaluation of keyphrase extraction algorithms. The platform currently encompasses 19 algorithms for automatic keyphrase extraction and methods for evaluating their performance against a diverse gold standard dataset. Among the 19 algorithms is a  keyphrase extraction method called BibRank. BibRank exploits an information-rich dataset created by parsing bibliographic data in BibTeX format. It combines a new weighting technique applied to the bibliographic data with positional, statistical, and word co-occurrence information.

The main contributions of this paper are as follows:

\begin{enumerate}
  \item BibRank dataset: Construction of an information-rich dataset by parsing publicly available bibliographic data, which includes manually assigned keywords.
  
  \item BibRank algorithm: Introduction of the BibRank algorithm, a novel method for keyphrase extraction that utilizes the bibliographic information within the BibRank dataset and statistical information.
  
  \item BibRank platform: Provision of a downloadable platform that integrates the BibRank dataset, BibRank algorithm, and other state-of-the-art keyphrase extraction algorithms. The platform includes evaluation metrics and allows for the integration of keyphrase extraction algorithms and datasets.
  
  \item Manual evaluation of keyphrases: Keyphrase extraction algorithms are evaluated using gold standard datasets as a benchmark. In our evaluation process, we rely on expert human evaluators to assess the quality and effectiveness of these gold-standard algorithms.
  
\end{enumerate}

The remaining sections of the paper closely align with the contributions presented earlier. The next section briefly overviews notable keyphrase extraction algorithms and datasets. Section \ref{BibRank} introduces the heterogeneous BibRank dataset and presents the BibRank algorithm. Section \ref{Results} concentrates on the automatic evaluation of the BibRank algorithm and other state-of-the-art algorithms. Moreover, this section includes assessing the gold standard algorithms' quality, guided by expert human evaluators. The paper concludes by summarizing our findings.

\section{Related work} \label{RW}

This section provides an overview of the essential stages in the automatic keyword extraction algorithms pipeline, highlighting the algorithms that influenced BibRank. 

The keyword extraction pipeline comprises linguistic preprocessing, candidate phrase selection, keyphrase feature selection, and keyphrase ranking and selection.
The text is segmented into sentences and tokenized into words during linguistic preprocessing. Several language processing techniques are applied, including lemmatization, stemming, POS tagging, stop word removal, and Named Entity Recognition (NER) \cite{7805062}. Sometimes, POS tagging is followed by syntactic parsing, and NER is particularly valuable in languages with reliable NER systems.
Candidate phrases are selected from the processed text using n-gram sequencing and noun-phrase chunking (NP chunking) \cite{mihalcea2004textrank}. Rules-based on acceptable sequences of POS tags, such as selecting sequences starting with adjectives and ending with a noun in English, are employed \cite{hasan2014automatic} to reduce the number of candidate phrases.

The subsequent step in the pipeline is feature selection for candidate phrases. Two types of features are calculated: in-document features and external features \cite{7805062}. In-document features can be statistical \cite{danesh2015sgrank}, positional \cite{AlamiMerrouni2019AutomaticKE} linguistic \cite{papagiannopoulou2020review} or context-based \cite{caragea-etal-2014-citation}. Statistical features like TF-IDF score are commonly used, while positional features indicate the candidate phrase's location in the title, abstract, or main text. Context features, such as sentence embeddings computed by deep neural networks, are also utilized. 
External features require resources like Wikipedia \cite{li2010semi} to quantify the association strengths between keyphrases. An example of a supervised keyphrase extraction algorithm that utilizes external features is CeKE \cite{caragea-etal-2014-citation}. CeKE employs citation-based features created from the references used in a publication. 

The assignment of weights to each candidate phrase is based on the calculated features in the keyphrase ranking and selection step. Subsequently, the candidate phrases are sorted, and the most relevant ones are selected using an experimental threshold. 

In the context of unsupervised methods, graph-based ranking algorithms like TextRank \cite{mihalcea2004textrank} deserve to be mentioned. These algorithms draw inspiration from the Google PageRank algorithm \cite{Page1999ThePC} and have demonstrated success in text summarization and keyword extraction. The text document is represented as a graph, where candidate phrases are nodes, and their relationships are edges. These relationships can be co-occurrence relations \cite{beliga2015overview}, syntactic dependencies \cite{mihalcea2004textrank}, or semantic relations \cite{li2010semi}.

In the keyphrase ranking step, an adapted PageRank algorithm is employed, which iterates until convergence on the graph representation of the text, ultimately selecting the top-ranked candidate phrases. Another algorithm in this family is PositionRank \cite{florescu2017positionrank}. Building upon the principles of TextRank, PositionRank introduces a bias towards frequently occurring candidate phrases that appear early in the document. It operates at the word level, transforming the text into a graph, applying a position-based PageRank algorithm, and extracting candidate phrases.

Other initiatives that share a connection with our work encompass the creation and visualization of bibliometric networks. VosViewer stands out as a notable tool in these endeavors \cite{vanEck2010}.
While VosViewer is not specifically a tool for keyphrase extraction, it is a relevant software used for creating and visualizing bibliometric networks. These networks can encompass journals, researchers, or single publications, helping to analyze and visualize trends and patterns in scientific literature. VosViewer provides multiple avenues to build, visualize, and investigate bibliometric networks, simplifying the process for users to gain insights from bibliometric data

\section{BibRank} \label{BibRank}

\subsection{BibRank Dataset} \label{BibRank dataset}

Keyphrase datasets serve as the standard for evaluating automatic keyphrase extraction methods, encompassing texts and lists of associated keyphrases. These gold standards are widely available across scientific publications, news articles, and web posts \cite{papagiannopoulou2020review}.

We utilize BibTeX entries from the web to construct a new and information-rich keyphrase extraction dataset. Unlike existing datasets that often include only the abstract, full article text, title, and keywords of a document, our dataset incorporates additional metadata such as the publication year, journal title, and author name. 
An example of a BibTeX record for a publication is illustrated in Figure \ref{fig:bibtex}, where the entry type (e.g., "Article") is indicated after the "@," followed by various attributes (e.g., author, title, journal, and paper keywords) and their respective values.

\begin{figure} [hbtp!]
\begin{lstlisting}[basicstyle=\footnotesize]
@Article{Wang:2009:EKF,
  author =       "Zidong Wang and Xiaohui Liu and Yurong Liu and Jinling
                 Liang and Veronica Vinciotti",
  title =        "An Extended {Kalman} Filtering Approach to Modeling
                 Nonlinear Dynamic Gene Regulatory Networks via Short
                 Gene Expression Time Series",
  journal =      j-TCBB,
  volume =       "6",
  number =       "3",
  pages =        "410--419",
  month =        jul,
  year =         "2009",
  CODEN =        "ITCBCY",
  DOI =          "https://doi.org/10.1109/TCBB.2009.5",
  ISSN =         "1545-5963 (print), 1557-9964 (electronic)",
  ISSN-L =       "1545-5963",
  bibdate =      "Tue Aug 11 18:13:22 MDT 2009",
  bibsource =    "http://portal.acm.org/;
                 http://www.math.utah.edu/pub/tex/bib/tcbb.bib",
  abstract =     "In this paper, the extended Kalman filter (EKF)
                 algorithm is applied to model the gene regulatory
                 network from gene time series data. The gene regulatory
                 network is considered as a nonlinear dynamic stochastic
                 model that consists of the gene measurement equation
                 and the gene regulation equation. After specifying the
                 model structure, we apply the EKF algorithm for
                 identifying both the model parameters and the actual
                 value of gene expression levels. It is shown that the
                 EKF algorithm is an online estimation algorithm that
                 can identify a large number of parameters (including
                 parameters of nonlinear functions) through iterative
                 procedure by using a small number of observations. Four
                 real-world gene expression data sets are employed to
                 demonstrate the effectiveness of the EKF algorithm, and
                 the obtained models are evaluated from the viewpoint of
                 bioinformatics.",
  acknowledgement = ack-nhfb,
  fjournal =     "IEEE/ACM Transactions on Computational Biology and
                 Bioinformatics",
  journal-URL =  "http://portal.acm.org/browse_dl.cfm?idx=J954",
  keywords =     "clustering; DNA microarray technology; extended Kalman
                 filtering; gene expression; Modeling; time series
                 data.",
}
\end{lstlisting}
\caption{BibTeX record example}
\label{fig:bibtex}
\end{figure}

Publicly available BibTeX records can be found in online archives like the TUG bibliography archive. TUG's archive contains a vast collection of over 1.6 million categorized BibTeX records from various journals. The archive supports search capabilities using SQL commands \cite{beebe2009bibtex}.

To create the BibRank dataset, we processed more than 30,000 BibTeX records extracted from the TUG bibliography archive. Currently, the dataset consists of 18,193 unique records with 22 attributes. These attributes represent the distinct values in all the bib records, including publication year, journal of publication, and bib archive. The dataset includes publications from 1974 to 2019. Table \ref{tab:bibrankdataset} provides statistics on authors, journals, topics, and bib files covered by the dataset.

The bib files, referring to the archives or databases from which the papers were imported, were categorized into one of the following 12 topics: science history journals, computer science journals and topics, ACM Transactions, cryptography, fonts and typography, IEEE journals, computational/quantum chemistry/physics, numerical analysis, probability and statistics, SIAM journals, mathematics, and mathematical and computational biology. Expanding the dataset by processing additional bibliography files in BibTeX format is possible.

The file for the dataset and the essential tools for altering and producing new datasets are available in the BibRank project's GitHub repository. This repository grants users access to the original data and equips them with the requisite resources for customizing the data to their particular requirements or generating entirely new datasets 
\begin{table}[h]
\centering
\caption{BibRank Dataset}
\begin{tabular}{| l | l | }
	\hline
	Data & Count \\
	\hline
	Records (abstracts) & 18,193  \\
	\hline
	Authors & 16,883 \\ 
	\hline
	Journals & 693  \\
	\hline
	Bib Files & 285 \\
	\hline
	Topics &  12 \\
	\hline
	Avg Words & 121 \\
	\hline
	Avg Keyphrases & 9 \\
	\hline
\end{tabular}
\label{tab:bibrankdataset}
\end{table}

\subsection {BibRank algorithm} \label{BibRank algorithm}

The BibRank algorithm, comprising five steps, presents an innovative method for weighting candidate phrases, emphasizing the abstracts of scientific publications and based on the concept of a context for a group of BibTeX records.

\begin{enumerate}

\item Candidate Selection. 
The candidate phrases in the document are noun chunks. To identify the noun chunks, we apply rules based on sequences of POS tags. In our workflow, we use the Stanford CoreNLP Natural Language Processing Toolkit \cite{manning2014stanford}, but other noun chunkers can be easily integrated into the platform.

\item PositionRank Weight Calculation. 
The PositionRank algorithm \cite{florescu2017positionrank} assigns position weights to candidate phrases. Higher weights are given to the words appearing earlier in the document. For example, if a phrase consists of positions 3, 6, and 8, its weight is calculated as follows: $\frac{1}{3}+\frac{1}{5}+\frac{1}{8}=\frac{5}{8}=0.625$. The final weight of each candidate phrase is determined by summing and normalizing the position weights of each word in the phrase. Additionally, the scores of each word are recursively computed using the PageRank algorithm, as described by Equation \ref{eqn:nbib1} \cite{florescu2017positionrank, mihalcea2004textrank}.

\begin{equation} \label{eqn:nbib1} 
S\left(v_{i}\right)=(1-d) \cdot \hat{p_{i}}+d \cdot \sum_{v_{j} \in \operatorname{In}\left(v_{i}\right)} \frac{w_{j i}}{Out\left(v_{j}\right)} S\left(v_{j}\right) 
\end{equation}

In Equation \ref{eqn:nbib1}, $S(v_i)$ represents the weight of each word $i$ in a candidate phrase $p$, represented by the vertex $v_i$. The damping factor $d$ reflects the probability of jumping to a random vertex in the graph, and $\hat{p}$ is the position weight of the word $i$. The set $\operatorname{In}(v_{i})$ contains the adjacent vertices pointing to vertex $i$, and $w_{ji}$ is the edge weight between $v_i$ and $v_j$. Finally, $Out(v_{j})$ is the set of adjacent vertices pointed to by vertex $i$, and is computed as $\sum_{V_{k} \in Out(V_{j})} w_{j k}$.
    
 \item Context Formulation.
The computation of the context for a publication involves selecting a set of BibTeX records according to specific criteria. For instance, if we consider a computer science article published in 2012, the context could be formed by including all computer science papers published within the same year. With the original BibRank dataset containing 22 attributes, each attribute can potentially define a distinct context.

\item Bib Weight Calculation.
The bib weights aim to capture the occurrence frequency of candidate phrases within the context. Each record includes a list of keyphrases, allowing for the calculation of weights for candidate phrases based on Equation \ref{bib_weight_calculation}.

    \begin{equation} \label{bib_weight_calculation}
    \lambda_p=\frac{1}{\alpha} \sum_{d \subseteq D} c_{p d}   
    \end{equation}

$\lambda{p}$ is the bib weight, $\alpha$ is a factor used for normalization, $D$ is the set of all records that belong to the chosen context, $d$ is a record, and $c$ is the occurrence of a candidate phrase in the record's keyphrases list. $\alpha$ was calculated as the maximum bib weight across all keyphrases in the context documents. 

\item Candidate Phrase Ranking and Selection.
The ranking of candidate phrases is determined by combining their bib weights and position scores. The scores of individual words within each candidate phrase are added to the phrase's bib weight, resulting in a sum that determines the final ranking of the candidate phrases, as illustrated in Equation \ref{candidate_phrase_ranking}. The document's keyphrases are then determined by selecting the top $N$ candidate phrases.

    \begin{equation} \label{candidate_phrase_ranking}
        S_{final}\left(p\right)= {\sum_{v_i \in \operatorname{V_p}} S(v_i)} +  \lambda_p
    \end{equation}
    
     $V_p$ is the set of words that belongs to candidate phrase $p$ and $\lambda_p$ is the calculated bib weight for the candidate phrase $p$. 
\end{enumerate}

 In the illustrated Figure \ref{fig:BibRank Keyphrases Extraction Example}, The BibRank algorithm begins by processing the input text, extracting nouns and noun phrases like 'Keyword' and 'automatic identification,' which are considered as selected candidates. It then infers keyphrases, including 'Keyword extraction' and 'automatic identification,' assigning them scores of 0.38 and 0.30, respectively. These scores denote their relevance and significance to the document's main topic, calculated based on position weight and Bib weights.

\begin{figure} [h]
\fbox{%
\begin{minipage} {\textwidth}
\begin{itemize}

	\item Input Text: Keyword extraction is tasked with the automatic identification of terms that best describe the subject of a document.
	\item  Nouns and Noun phrases: 'extraction', 'subject', 'Keyword', 'identification', 'automatic identification', 'document', 'terms', 'Keyword extraction', 'best', 'automatic'
	\item  Keyphrases: 'Keyword extraction', 'automatic identification', 'extraction', 'automatic', 'Keyword', 'identification'
        \item  scores: 0.38, 0.30, 0.23, 0.17, 0.15, 0.13
\end{itemize}
\end{minipage}
}
\caption{BibRank Keyphrases Extraction Example}
\label{fig:BibRank Keyphrases Extraction Example}
\end{figure}

\subsection {BibRank platform} \label{BibRank platform}

BibRank is a versatile online platform developed in Python that simplifies the integration of keyphrase extraction algorithms, encompassing three modules: Datasets, Algorithms, and Evaluation.

One of the standout attributes of the platform is its comprehensive support for keyphrase extraction datasets. It seamlessly incorporates user datasets and features multiple pre-integrated datasets, such as the BibRank dataset (see \ref{BibRank dataset}) and five others extensively detailed in table \ref{tab:datasets}. This table provides crucial information about the papers linked to each dataset, the number of documents contained, and the document types, distinguishing between abstracts and full papers.

Moreover, BibRank facilitates users in crafting personalized datasets with ease. The platform offers user-friendly routines tailored to process BibTeX files, simplifying the generation of new datasets that align with the user's specific needs and requirements.

\begin{table}[h]
\centering
\caption{BibRank platform Datasets}
\begin{tabular}{| c | c | c |}
	\hline
	Dataset & Documents& Type  \\
	\hline
	ACM \cite{schutz2008keyphrase} & 2,304 & Full papers  \\
	\hline
	NUS \cite{nguyen2007keyphrase} & 211 & Full papers  \\
	\hline
	Inspec \cite{hulth2003improved} & 2,000  & Abstracts \\
	\hline
	WWW \cite{caragea-etal-2014-citation}  & 1,330  & Abstracts \\
	\hline
	KDD \cite{caragea-etal-2014-citation} & 755  & Abstracts \\
	\hline
	BibRank Dataset & 18,193  & Abstracts and Metadata \\
	\hline
\end{tabular}
\label{tab:datasets}
\end{table}

The platform offers a comprehensive range of keyphrase extraction algorithms, including the BibRank algorithm (refer to \ref{BibRank algorithm}) and ten additional ones, all clearly specified in table \ref{tab:models}. It provides a user-friendly interface for effortlessly integrating the user's own keyphrase extraction algorithms. For smooth integration, the user's algorithm must extend a superclass that encompasses the blueprint for the crucial extraction operations, where the algorithm's name is designated as a class attribute. Additionally, the algorithm must incorporate a function that efficiently returns the extracted keyphrases and their corresponding weights. The platform incorporates PKE, an open-source toolkit for keyphrase. \cite{boudin:2016:COLINGDEMO}.

\begin{table}[h]
\centering
\caption{BibRank platform Models}
\begin{tabular}{| c | c | c | c |}
	\hline
	Method & Year & Approach Type \\
	\hline
	TFIDF \cite{frank1999domain}  & 1999 &  Statistical \\
	\hline
	KPMiner \cite{el2010kp}  & 2010 &  Statistical\\
	\hline
	YAKE \cite{campos2020yake}  & 2020 &  Statistical  \\
	\hline
	TextRank \cite{mihalcea2004textrank} & 2004 & Graph based \\
	\hline
	CollabRank \cite{wan2008collabrank} & 2008 & Graph based \\
	\hline
	TopicRank \cite{bougouin2013topicrank} & 2013 & Graph based \\
	\hline
	PositionRank \cite{florescu2017positionrank}  & 2017 & Graph based \\
	\hline
	SGRank \cite{danesh2015sgrank}  & 2015 & Hybrid Statistical-graphical \\
	\hline
	sCAKE \cite{duari2019scake}  & 2018 & Hybrid Statistical-graphical  \\
	\hline
	KeyBERT \cite{zenodo} & 2021 & Sentence Embeddings \\
	\hline
\end{tabular}
\label{tab:models}
\end{table}

To assess the accuracy of a keyphrase extraction algorithm on a given dataset, the platform provides an evaluation module in the form of a Python script. Users can select the algorithm to be evaluated and specify the metadata for the dataset, such as the year of publication or journal. The evaluation script computes the recall (R), precision (P), and F1 scores, widely recognized as standard measures of algorithm performance.

\section {Results} \label{Results}

\subsection{Evaluation methodology}

The widely accepted assumption that the gold standard serves as the reference truth for evaluating algorithms is acknowledged. However, a comprehensive twofold evaluation process was conducted to examine this assumption critically. The first evaluation aimed to assess the algorithms against the gold standard, while the second evaluation focused on evaluating the gold standard itself.

Datasets with manually assigned keywords were used as benchmarks to assess the algorithms' performance. The evaluations were carried out using the BibRank platform, where the algorithms were tested on the BibRank dataset with parameter adjustments. The default setting for the first parameter, determining the number of keywords to extract, was $10$ for all algorithms. The second parameter, the tokenizer, utilized the Stanford CoreNLP toolkit, as explained in the BibRank algorithm section. The damping factor $\alpha$ was set to $0.85$, and the window size was set to $2$ based on experiments by \cite{florescu2017positionrank}. Extracted keyphrases were compared to the manually assigned keywords in the gold standard dataset to measure the algorithms' performance, considering exact matches as successful hits. Standard evaluation metrics such as recall, precision, and F1 score were computed.

Evaluators with expertise were sought through a reputable freelancing platform to evaluate the gold standard. These evaluators were carefully selected based on specific criteria, including fluency in English and a proven track record in similar tasks. Two experts were assigned to evaluate 100 annotated documents containing keywords using seven algorithms and the gold standard. The evaluators were kept unaware of the algorithm names or the gold standard during the evaluation process to prevent potential bias. The evaluators meticulously annotated the different data sets using a five-point scale:

\begin{enumerate}

\item Very bad: The keywords are considered inadequate and do not meaningfully represent the text.

\item Bad: The keywords are a mix of poor and good choices, lacking consistency and not fully capturing the essence of the text.

\item Acceptable: The keywords are generally satisfactory and represent the text to a reasonable extent.

\item Good: The keywords are of good quality, although they may not fully encompass all the text's main ideas.

\item Very good: The provided keywords accurately summarize the text and effectively capture the main ideas.

\end{enumerate}

Overall, our twofold evaluation approach provides a comprehensive analysis of both the algorithm and the gold standard, allowing us to understand the strengths and weaknesses of each.

\subsection{Results}
  
The evaluation of the algorithms involved three experiments, each utilizing a different section of the BibRank dataset. The experiments focused on specific domains, namely "Computer science (compsci)," "ACM," and "history, philosophy, and science," consisting of 335, 127, and 410 papers, respectively. In choosing the dataset years, we aimed for diverse temporal coverage and ran tests on various combinations to ensure validity. For Computer science (compsci), bib scores were generated using publications from the years 1980 to 1987, and the test data was sourced from publications in 1988; ACM bib scores were derived from 1990 to 1996 and tested against 1997 to 2020 publications; for "history, philosophy, and science," scores were based on 2009 to 2011, testing with 2012 to 2014 publications. For a comprehensive overview of these experiments, including the categories used, please refer to Table \ref{tab:Full results}. The table displays the categories the articles belong to and seven selected algorithms for evaluation. We selected these algorithms to exemplify various keyphrase extraction approaches discussed in the Related Works section, showcasing the implementation of distinct methodologies for keyword extraction.

Upon closer inspection,  the BibRank algorithm demonstrates consistent enhancements across different datasets, as can be seen in the tables \ref{tab:bibrank improv compsci}, \ref{tab:bibrank improv science}, and \ref{tab:bibrank improv probstat}. When compared to TextRank and PositionRank, which use comparable techniques, the integration of Bib Weights in the BibRank algorithm leads to a noticeable enhancement in performance.

\begin{enumerate}
 
 \item YAKE (Yet Another Keyword Extractor) is a statistical keyphrase extraction algorithm that utilizes a "maximal marginal relevance" approach to promote diversity in the selected keywords. This ensures that the extracted keyphrases cover a wide range of topics and concepts.
 
 \item The SGRank and sCake methods are algorithms used to extract keyphrases from a document. They employ statistical analysis and graph-based techniques, blending both advantages to identify important keywords. Notably, sCake stands out for integrating domain-specific knowledge into its process when analyzing documents.

 \item {KeyBERT} represents a user-friendly and lightweight algorithm for keyword extraction. It harnesses the power of BERT transformers' embeddings to identify important keywords in a given text. Using an unsupervised technique, KeyBERT calculates the cosine similarity between each phrase and document to determine the most relevant keyphrases.

\end{enumerate}

The preceding sections contain in-depth discussions about graph-based techniques, including TextRank, PositionRank, and BibRank. These algorithms use graph-based approaches to analyze word relationships and extract essential keywords from a text.

Our objective in incorporating these algorithms is to comprehensively evaluate various keyphrase extraction techniques.

\begin{figure}[ht]
    \centering
    \includegraphics[width=0.8\textwidth]{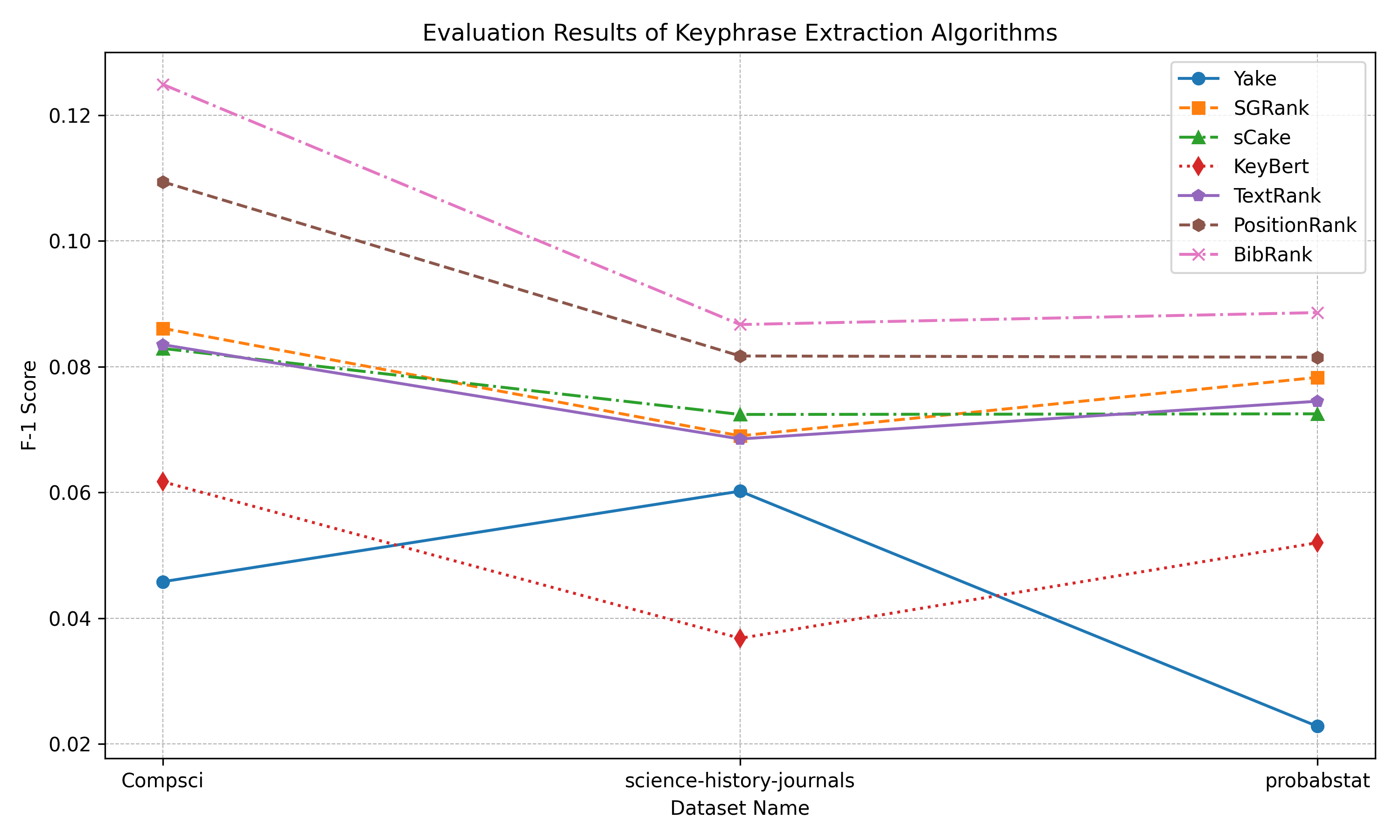}
    \caption{Evaluation Results of Keyphrase Extraction Algorithms.}
    \label{fig:keyphrase_evaluation}
\end{figure}

In addition to using standard gold keyphrases, the chosen experts manually evaluated seven keyphrase extraction approaches. To gauge the performance of each method, the experts assigned scores from 1 to 5 to the generated keywords for 100 randomly selected documents. Table \ref{tab:manual_eval} summarizes the average performance of each evaluated approach. These evaluations offer valuable insights into the effectiveness of the diverse keyphrase extraction methods.
The figure denoted by \ref{fig:keyphrase_evaluation} provides a clear and organized visual display of the results for the keyphrase extraction algorithms. These algorithms were evaluated based on the domains depicted on the x-axis, while the F1 score is plotted on the y-axis.

  \begin{table}[]
  \caption{Evaluation Results of selected keyphrase extraction algorithms, including BibRank}
  \resizebox{\textwidth}{!}{
\begin{tabular}{|l|lll||lll||lll|}
\hline
                      & \multicolumn{3}{c||}{Compsci}                                                                          & \multicolumn{3}{c||}{science-history-journals}                                                         & \multicolumn{3}{c|}{Probstat}                                                                         \\ \hline
                      & \multicolumn{1}{c|}P      & \multicolumn{1}{c|}{R}      & \multicolumn{1}{c||}{F1} & \multicolumn{1}{c|}{P}      & \multicolumn{1}{c|}{R}      & \multicolumn{1}{c||}{F1} & \multicolumn{1}{c|}{P}      & \multicolumn{1}{c|}{R}      & \multicolumn{1}{c|}{F1} \\ \hline
Yake         & \multicolumn{1}{l|}{0.0728}          & \multicolumn{1}{l|}{0.0367}          & 0.0458                           & \multicolumn{1}{l|}{0.0606}          & \multicolumn{1}{l|}{0.0705}          & 0.0602                           & \multicolumn{1}{l|}{0.0171}          & \multicolumn{1}{l|}{0.0366}          & 0.0228                           \\ \hline
SGRank       & \multicolumn{1}{l|}{0.1282}          & \multicolumn{1}{l|}{0.0730}          & 0.0861                           & \multicolumn{1}{l|}{0.0645}          & \multicolumn{1}{l|}{0.0903}          & 0.0690                           & \multicolumn{1}{l|}{0.0594}          & \multicolumn{1}{l|}{0.1235}          & 0.0783                           \\ \hline
sCake        & \multicolumn{1}{l|}{0.1213}          & \multicolumn{1}{l|}{0.0714}          & 0.0829                           & \multicolumn{1}{l|}{0.0676}          & \multicolumn{1}{l|}{0.0949}          & 0.0724                           & \multicolumn{1}{l|}{0.0549}          & \multicolumn{1}{l|}{0.1141}          & 0.0725                           \\ \hline
KeyBert      & \multicolumn{1}{l|}{0.0839}          & \multicolumn{1}{l|}{0.0564}          & 0.0617                           & \multicolumn{1}{l|}{0.0315}          & \multicolumn{1}{l|}{0.0501}          & 0.0368                           & \multicolumn{1}{l|}{0.0380}          & \multicolumn{1}{l|}{0.0880}          & 0.0520                           \\ \hline
TextRank     & \multicolumn{1}{l|}{0.1236}          & \multicolumn{1}{l|}{0.0716}          & 0.0835                           & \multicolumn{1}{l|}{0.0621}          & \multicolumn{1}{l|}{0.0912}          & 0.0685                           & \multicolumn{1}{l|}{0.0562}          & \multicolumn{1}{l|}{0.1175}          & 0.0745                           \\ \hline
PositionRank & \multicolumn{1}{l|}{0.1579}          & \multicolumn{1}{l|}{0.0953}          & 0.1094                           & \multicolumn{1}{l|}{0.0740}          & \multicolumn{1}{l|}{0.1102}          & 0.0817                           & \multicolumn{1}{l|}{0.0605}          & \multicolumn{1}{l|}{0.1347}          & 0.0815                           \\ \hline
BibRank      & \multicolumn{1}{l|}{0.1812} & \multicolumn{1}{l|}{0.109} & {0.1249}                  & \multicolumn{1}{l|}{0.0811} & \multicolumn{1}{l|}{0.1136} & 0.0867                  & \multicolumn{1}{l|}{0.0659} & \multicolumn{1}{l|}{0.1457} & 0.0886                  \\ \hline
\end{tabular}
}
\label{tab:Full results}
\end{table}

\begin{table}[]
\centering
\caption{BibRank Improvements: Compsci}
\begin{tabular}{|c|c|l|l|l|}
\hline
\multicolumn{1}{|l|}{}                      & \multicolumn{1}{l|}{Bib Weights Records} & \multicolumn{1}{c|}{P} & \multicolumn{1}{c|}{R} & \multicolumn{1}{c|}{F1} \\ \hline
\multicolumn{1}{|l|}{TextRank}     & 0                                                 & 0.1236                          & 0.0716                          & 0.0835                           \\ \hline
\multicolumn{1}{|l|}{PositionRank} & 0                                                 & 0.1579                          & 0.0953                          & 0.1094                           \\ \hline
\multirow{4}{*}{BibRank}           & 299                                               & 0.1764                          & 0.1065                          & 0.1216                           \\ \cline{2-5} 
                                            & 976                                               & 0.1764                          & 0.1063                          & 0.1218                           \\ \cline{2-5} 
                                            & 1155                                              & 0.1809                          & 0.1083                          & 0.1242                           \\ \cline{2-5} 
                                            & 1746                                              & 0.1812        & 0.109       & 
                                            0.1249         \\ \hline
\end{tabular}
\label{tab:bibrank improv compsci}
\end{table}

\begin{table}[]
\centering
\caption{BibRank Improvements: science-history-journals}
\begin{tabular}{|l|clll|}
\hline
                                  & \multicolumn{4}{c|}{Science-history-journals}                                                                                                                       \\ \hline
                                  & \multicolumn{1}{l|}{Bib Weights} & \multicolumn{1}{c|}{P}               & \multicolumn{1}{c|}{R}               & \multicolumn{1}{c|}{F1} \\ \hline
TextRank                          & \multicolumn{1}{c|}{0}          & \multicolumn{1}{l|}{0.0621}          & \multicolumn{1}{l|}{0.0912}          & 0.0685                           \\ \hline
PositionRank                      & \multicolumn{1}{c|}{0}          & \multicolumn{1}{l|}{0.0740}          & \multicolumn{1}{l|}{0.1102}          & 0.0817                           \\ \hline
\multirow{3}{*}{BibRank}          & \multicolumn{1}{c|}{54}         & \multicolumn{1}{l|}{0.0780}          & \multicolumn{1}{l|}{0.1098}          & 0.0833                           \\ \cline{2-5} 
                                  & \multicolumn{1}{c|}{81}         & \multicolumn{1}{l|}{0.0787}          & \multicolumn{1}{l|}{0.1108}          & 0.0842                           \\ \cline{2-5} 
                                  & \multicolumn{1}{c|}{173}        & \multicolumn{1}{l|}{0.0811} & \multicolumn{1}{l|}{0.1136} & 0.0867                  \\ \hline
\end{tabular}
\label{tab:bibrank improv science}
\end{table}

\begin{table}[]
\centering
\caption{BibRank Improvements: probstat}
\begin{tabular}{|l|c|l|l|l|}
\hline
                                                        & \multicolumn{1}{l|}{Bib Weights Records} & P       & R       & F1      \\ \hline
TextRank                                                & 0                                       & 0.0562  & 0.1175  & 0.0745  \\ \hline
PositionRank                                            & 0                                       & 0.0605  & 0.1347  & 0.0815  \\ \hline
\multicolumn{1}{|c|}{\multirow{3}{*}{BibRank}}          & 139                                     & 0.0646  & 0.1422  & 0.0868  \\ \cline{2-5} 
\multicolumn{1}{|c|}{}                                  & 237                                     & 0.0649  & 0.1423  & 0.0870  \\ \cline{2-5} 
\multicolumn{1}{|c|}{}                                  & 367                                     & 0.0659  & 0.1457  & 0.0886   \\ \hline
\end{tabular}
\label{tab:bibrank improv probstat}
\end{table}

\begin{table}[]
\caption{Manual Evaluation}
\centering
\begin{tabular}{|l|l|l|}
\hline
Model         & Expert 1 & Expert 2 \\ \hline
Gold Standard & 2.85     & 2.08     \\ \hline
Yake          & 2.95     & 1.65     \\ \hline
SGRank        & 4.2      & 3.47     \\ \hline
sCake         & 3.71     & 3.39     \\ \hline
KeyBert       & 4.61     & 4.39     \\ \hline
TextRank      & 4.77     & 3.99     \\ \hline
PositionRank  & 4.41     & 3.37     \\ \hline
BibRank       & 4.4      & 3.77     \\ \hline
\end{tabular}
\label{tab:manual_eval}
\end{table}

\subsection{Discussion}

The Yake algorithm and the gold standard sets of keyphrases received the lowest scores from the experts in our evaluation. This result was expected for Yake, as it is the only statistical approach among the evaluated techniques. Prior research \cite{hasan2014automatic} has also indicated that models relying on statistical features exhibit lower average performance in keyphrase extraction tasks. However, the surprising finding was the performance of the gold standard keyphrases.

We conducted interviews with the experts who participated in the evaluation to gain deeper insights. One expert mentioned that the gold standard keyphrases are overly general and limited in scope. They are designed to capture the central ideas or keyphrases of the document, which may result in the omission of some important keywords. In contrast, algorithms such as BibRank, PositionRank, TextRank, and KeyBERT better understood the document's meaning, enabling them to extract more relevant and specific keyphrases.

Figure \ref{fig:Keywords Example}  presents an abstract that the experts evaluated, and the corresponding scores provided by the experts are listed in table \ref{tab:manual_eval_fig2}. The gold standard keywords received low scores despite including important keyphrases like "Chinese dependency parsing" and "unlabeled data." However, there were cases where essential keyphrases were missing, while some keywords not explicitly mentioned in the abstract were included in the gold standard set. For instance, the term "semi-supervised learning" was incorporated in the gold standard keyword list but did not appear in the original abstract.

Yake achieved a low score, indicating that the algorithm lacks the contextual understanding exhibited by the other keyword extraction methods.

SGRank outperformed the gold standard, effectively highlighting essential keywords such as "long-distance word," "unlabeled attachment score," and "supervised learning method."

SCake also demonstrated strong performance, successfully extracting detailed keywords related to different types of dependency parsers and incorporating "short dependency information."

KeyBERT showcased robust performance, extracting comprehensive keywords such as "improves parsing performance" and "parsing approach incorporating," which enhanced the understanding of the paper's content.

TextRank consistently performed well, generating similar keywords to SCake and SGRank, indicating its consistency in identifying key concepts.

PositionRank, with a score of 5, provided additional context by introducing terms such as "short dependencies."

BibRank consistently scored 5 in both evaluations, effectively extracting keywords related to various parser types, "short dependency information," and specific performance metrics like "high performance." It also included additional contextual keywords, such as "machine translation," providing a comprehensive overview of the abstract's content.

Overall, these evaluations shed light on the strengths and weaknesses of different keyphrase extraction methods and help us understand their performance characteristics in the context of academic literature.

The detailed results of our evaluations, substantiating the findings discussed in this paper, are recorded and made available for public scrutiny and exploration. These results can be found in our GitHub repository's "evaluation\_results" folder.

\begin{table}[]
\caption{The expert evaluation for the abstract presented in figure \ref{fig:Keywords Example}}
\centering
\begin{tabular}{|l|l|l|}
\hline
Model         & Expert 1 & Expert 2 \\ \hline
Gold Standard & 2        & 1        \\ \hline
Yake          & 3        & 1        \\ \hline
SGRank        & 4        & 3        \\ \hline
sCake         & 4        & 3        \\ \hline
KeyBert       & 4        & 4        \\ \hline
TextRank      & 5        & 4        \\ \hline
PositionRank  & 5        & 5        \\ \hline
BibRank       & 5        & 5        \\ \hline
\end{tabular}
\label{tab:manual_eval_fig2}
\end{table}

\begin{figure} [h]
\fbox{%
\begin{minipage} {\textwidth}
\begin{itemize}

	\item Abstract: dependency parsing has become increasingly popular for a surge of interest lately for applications such as machine translation and question answering. currently, several supervised learning methods can be used for training high-performance dependency parsers if sufficient labeled data are available. however, currently used statistical dependency parsers provide poor results for words separated by long distances. in order to solve this problem, this article presents an effective dependency parsing approach of incorporating short dependency information from unlabeled data. the unlabeled data is automatically parsed by using a deterministic dependency parser, which exhibits a relatively high performance for short dependencies between words. we then train another parser that uses the information on short dependency relations extracted from the output of the first parser. the proposed approach achieves an unlabeled attachment score of 86.52\%, an absolute 1.24\% improvement over the baseline system on the chinese treebank data set. the results indicate that the proposed approach improves the parsing performance for longer distance words.
	\item gold: 'chinese dependency parsing', 'semi-supervised learning', 'unlabeled data'
    \item Yake: 'dependency', 'parser', 'datum', 'performance', 'word', 'unlabeled', 'short', 'approach', 'distance', 'high'
    \item Sgrank: 'long distance word', 'chinese treebank datum', 'unlabeled attachment score', 'deterministic dependency parser', 'short dependency relation', 'statistical dependency parser', 'short dependency information', 'performance dependency parser', 'supervised learning method', 'unlabeled datum'
	\item sCake: 'performance dependency parser', 'statistical dependency parser', 'deterministic dependency parser', 'short dependency information', 'dependency parsing', 'short dependency relation', 'effective dependency', 'unlabeled datum', 'chinese treebank datum', 'high performance'
    \item KeyBert: 'improves parsing performance', 'effective dependency parsing', 'performance dependency parsers', 'dependency parsing approach', 'dependency parsers provide', 'statistical dependency parsers', 'parsers provide poor', 'parsing performance', 'deterministic dependency parser', 'parsing approach incorporating'
    \item TextRank: 'performance dependency parser', 'deterministic dependency parser', 'statistical dependency parser', 'short dependency information', 'short dependency relation', 'dependency parsing', 'effective dependency', 'unlabeled datum', 'long distance word', 'chinese treebank datum'
    \item PositionRank: 'performance dependency parsers', 'statistical dependency parsers', 'deterministic dependency parser', 'short dependency information', 'short dependency relations', 'short dependencies', 'effective dependency', 'dependency', 'first parser', 'machine translation'
    \item BibRank: 'performance dependency parsers', 'statistical dependency parsers', 'deterministic dependency parser', 'short dependency information', 'short dependency relations', 'short dependencies', 'effective dependency', 'dependency', 'high performance', 'chinese treebank data'

\end{itemize}
\end{minipage}
}
\caption{Generative Model-based Keyphrase Extraction Example}
\label{fig:Keywords Example}
\end{figure}


\section{Conclusions} \label{Conclusions}

This paper introduces the BibRank platform, a versatile online platform developed in Python, which simplifies the integration of keyphrase extraction algorithms. A new keyphrase extraction dataset, the BibRank dataset, is presented to benchmark keyphrase extraction algorithms. The paper also introduces a state-of-the-art keyphrase extraction algorithm, BibRank, which utilizes the notion of context to compute keyphrases.

The main keyphrase extraction algorithms are comprehensively evaluated in the study using a two-fold approach: evaluating the algorithms against the gold standard and evaluating the gold standard itself. The evaluations are conducted on the BibRank dataset using standard evaluation metrics. Expert evaluators assess the gold standard using a five-point scale. The results demonstrate that some algorithms, such as BibRank and PositionRank, outperform the gold standard in extracting relevant and specific keyphrases, while others, like Yake, achieve lower scores due to their statistical nature. This evaluation provides valuable insights into the strengths and weaknesses of different keyphrase extraction methods in the context of academic literature.

The BibRank algorithm demonstrates state-of-the-art performance when evaluated against the gold standard. The authors encourage researchers to use the BibRank platform for evaluating their own keyphrase extraction algorithms. To ensure reproducibility, the BibRank platform, BibRank algorithm, and the BibRank dataset are publicly available (see the Data Availability Statement) for use by the research community.
Platforms such as BibRank and other keyphrase extraction tools have the potential to operate alongside VosViewer. If the research community starts using BibRank, we'll think about adding a plugin for integration with VosViewer.

\section{Data Availability} \label{data_availability}

The BibRank keyphrase extraction framework is readily available on GitHub to facilitate reproducibility. The repository includes:
\begin{itemize}
  \item The implementation of BibRank and 18 other keyphrase extraction methods.
  \item A detailed installation guide.
  \item Examples of evaluations.
  \item The Bib dataset used for evaluation.
  \item Comprehensive instructions for running experiments with the BibRank model.
  \item Reviewers full evaluation results. 
\end{itemize}

GitHub repository available at: \url{https://github.com/dallal9/Bibrank} (Accessed: 3 October 2023)

\section {Funding}
Eduard Barbu has been supported by the EKTB55 project "Teksti lihtsustamine eesti keeles"

\bibliographystyle{unsrtnat}
\bibliography{BibRank}

\begin{thebibliography}{26}
\providecommand{\natexlab}[1]{#1}
\providecommand{\url}[1]{\texttt{#1}}
\expandafter\ifx\csname urlstyle\endcsname\relax
  \providecommand{\doi}[1]{doi: #1}\else
  \providecommand{\doi}{doi: \begingroup \urlstyle{rm}\Url}\fi

\bibitem[Gusenbauer(2019)]{10.1007/s11192-018-2958-5}
Michael Gusenbauer.
\newblock Google scholar to overshadow them all? comparing the sizes of 12 academic search engines and bibliographic databases.
\newblock \emph{Scientometrics}, 118\penalty0 (1):\penalty0 177–214, jan 2019.
\newblock ISSN 0138-9130.
\newblock \doi{10.1007/s11192-018-2958-5}.
\newblock URL \url{https://doi.org/10.1007/s11192-018-2958-5}.

\bibitem[Merrouni et~al.(2016)Merrouni, Frikh, and Ouhbi]{7805062}
Zakariae~Alami Merrouni, Bouchra Frikh, and Brahim Ouhbi.
\newblock Automatic keyphrase extraction: An overview of the state of the art.
\newblock In \emph{2016 4th IEEE International Colloquium on Information Science and Technology (CiSt)}, pages 306--313, 2016.
\newblock \doi{10.1109/CIST.2016.7805062}.

\bibitem[Merrouni et~al.(2019)Merrouni, Frikh, and Ouhbi]{AlamiMerrouni2019AutomaticKE}
Zakariae~Alami Merrouni, Bouchra Frikh, and Brahim Ouhbi.
\newblock Automatic keyphrase extraction: a survey and trends.
\newblock \emph{Journal of Intelligent Information Systems}, 54:\penalty0 391 -- 424, 2019.

\bibitem[Mihalcea and Tarau(2004)]{mihalcea2004textrank}
Rada Mihalcea and Paul Tarau.
\newblock Textrank: Bringing order into text.
\newblock In \emph{Proceedings of the 2004 conference on empirical methods in natural language processing}, pages 404--411, 2004.

\bibitem[Hasan and Ng(2014)]{hasan2014automatic}
Kazi~Saidul Hasan and Vincent Ng.
\newblock Automatic keyphrase extraction: A survey of the state of the art.
\newblock In \emph{Proceedings of the 52nd Annual Meeting of the Association for Computational Linguistics (Volume 1: Long Papers)}, pages 1262--1273, 2014.

\bibitem[Danesh et~al.(2015)Danesh, Sumner, and Martin]{danesh2015sgrank}
Soheil Danesh, Tamara Sumner, and James~H Martin.
\newblock Sgrank: Combining statistical and graphical methods to improve the state of the art in unsupervised keyphrase extraction.
\newblock In \emph{Proceedings of the fourth joint conference on lexical and computational semantics}, pages 117--126, 2015.

\bibitem[Papagiannopoulou and Tsoumakas(2020)]{papagiannopoulou2020review}
Eirini Papagiannopoulou and Grigorios Tsoumakas.
\newblock A review of keyphrase extraction.
\newblock \emph{Wiley Interdisciplinary Reviews: Data Mining and Knowledge Discovery}, 10\penalty0 (2):\penalty0 e1339, 2020.

\bibitem[Caragea et~al.(2014)Caragea, Bulgarov, Godea, and Das~Gollapalli]{caragea-etal-2014-citation}
Cornelia Caragea, Florin~Adrian Bulgarov, Andreea Godea, and Sujatha Das~Gollapalli.
\newblock Citation-enhanced keyphrase extraction from research papers: A supervised approach.
\newblock In \emph{Proceedings of the 2014 Conference on Empirical Methods in Natural Language Processing ({EMNLP})}, pages 1435--1446, Doha, Qatar, October 2014. Association for Computational Linguistics.
\newblock \doi{10.3115/v1/D14-1150}.
\newblock URL \url{https://aclanthology.org/D14-1150}.

\bibitem[Li et~al.(2010)Li, Li, Li, Wang, and Qu]{li2010semi}
Decong Li, Sujian Li, Wenjie Li, Wei Wang, and Weiguang Qu.
\newblock A semi-supervised key phrase extraction approach: learning from title phrases through a document semantic network.
\newblock In \emph{Proceedings of the ACL 2010 conference short papers}, pages 296--300, 2010.

\bibitem[Page et~al.(1999)Page, Brin, Motwani, and Winograd]{Page1999ThePC}
Lawrence Page, Sergey Brin, Rajeev Motwani, and Terry Winograd.
\newblock The pagerank citation ranking : Bringing order to the web.
\newblock In \emph{The Web Conference}, 1999.

\bibitem[Beliga et~al.(2015)Beliga, Me{\v{s}}trovi{\'c}, and Martin{\v{c}}i{\'c}-Ip{\v{s}}i{\'c}]{beliga2015overview}
Slobodan Beliga, Ana Me{\v{s}}trovi{\'c}, and Sanda Martin{\v{c}}i{\'c}-Ip{\v{s}}i{\'c}.
\newblock An overview of graph-based keyword extraction methods and approaches.
\newblock \emph{Journal of information and organizational sciences}, 39\penalty0 (1):\penalty0 1--20, 2015.

\bibitem[Florescu and Caragea(2017)]{florescu2017positionrank}
Corina Florescu and Cornelia Caragea.
\newblock Positionrank: An unsupervised approach to keyphrase extraction from scholarly documents.
\newblock In \emph{Proceedings of the 55th Annual Meeting of the Association for Computational Linguistics (Volume 1: Long Papers)}, pages 1105--1115, 2017.

\bibitem[van Eck et~al.(2010)van Eck, Waltman, Dekker, and van~den Berg]{vanEck2010}
Nees~Jan van Eck, Ludo Waltman, Rommert Dekker, and Jan van~den Berg.
\newblock A comparison of two techniques for bibliometric mapping: Multidimensional scaling and {VOS}.
\newblock \emph{Journal of the American Society for Information Science and Technology}, 61\penalty0 (12):\penalty0 2405--2416, 2010.
\newblock ISSN 1532-2890.
\newblock \doi{10.1002/asi.21421}.
\newblock URL \url{http://dx.doi.org/10.1002/asi.21421}.

\bibitem[Beebe(2009)]{beebe2009bibtex}
Nelson~HF Beebe.
\newblock Bibtex meets relational databases.
\newblock \emph{j TUGboat}, 30:\penalty0 252--271, 2009.

\bibitem[Manning et~al.(2014)Manning, Surdeanu, Bauer, Finkel, Bethard, and McClosky]{manning2014stanford}
Christopher~D Manning, Mihai Surdeanu, John Bauer, Jenny~Rose Finkel, Steven Bethard, and David McClosky.
\newblock The stanford corenlp natural language processing toolkit.
\newblock In \emph{Proceedings of 52nd annual meeting of the association for computational linguistics: system demonstrations}, pages 55--60, 2014.

\bibitem[Schutz(2008)]{schutz2008keyphrase}
Alexander Schutz.
\newblock Keyphrase extraction from single documents in the open domain exploiting linguistic and statistical methods.
\newblock 2008.
\newblock URL \url{https://api.semanticscholar.org/CorpusID:8314070}.

\bibitem[Nguyen and Kan(2007)]{nguyen2007keyphrase}
Thuy~Dung Nguyen and Min-Yen Kan.
\newblock Keyphrase extraction in scientific publications.
\newblock In \emph{International conference on Asian digital libraries}, pages 317--326. Springer, 2007.

\bibitem[Hulth(2003)]{hulth2003improved}
Anette Hulth.
\newblock Improved automatic keyword extraction given more linguistic knowledge.
\newblock In \emph{Proceedings of the 2003 conference on Empirical methods in natural language processing}, pages 216--223, 2003.

\bibitem[Boudin(2016)]{boudin:2016:COLINGDEMO}
Florian Boudin.
\newblock pke: an open source python-based keyphrase extraction toolkit.
\newblock In \emph{Proceedings of COLING 2016, the 26th International Conference on Computational Linguistics: System Demonstrations}, pages 69--73, Osaka, Japan, December 2016.
\newblock URL \url{http://aclweb.org/anthology/C16-2015}.

\bibitem[FRANK(1999)]{frank1999domain}
E~FRANK.
\newblock Domain-specific keyphrase extraction.
\newblock In \emph{Proceedings of the 16th International Joint Conference on Artificial Intelligence, 1999}, pages 668--673, 1999.

\bibitem[El-Beltagy and Rafea(2010)]{el2010kp}
Samhaa~R El-Beltagy and Ahmed Rafea.
\newblock Kp-miner: Participation in semeval-2.
\newblock In \emph{Proceedings of the 5th international workshop on semantic evaluation}, pages 190--193, 2010.

\bibitem[Campos et~al.(2020)Campos, Mangaravite, Pasquali, Jorge, Nunes, and Jatowt]{campos2020yake}
Ricardo Campos, V{\'\i}tor Mangaravite, Arian Pasquali, Alipio Jorge, C{\'e}lia Nunes, and Adam Jatowt.
\newblock Yake! keyword extraction from single documents using multiple local features.
\newblock \emph{Information Sciences}, 509:\penalty0 257--289, 2020.

\bibitem[Wan and Xiao(2008)]{wan2008collabrank}
Xiaojun Wan and Jianguo Xiao.
\newblock Collabrank: towards a collaborative approach to single-document keyphrase extraction.
\newblock In \emph{Proceedings of the 22nd International Conference on Computational Linguistics (Coling 2008)}, pages 969--976, 2008.

\bibitem[Bougouin et~al.(2013)Bougouin, Boudin, and Daille]{bougouin2013topicrank}
Adrien Bougouin, Florian Boudin, and B{\'e}atrice Daille.
\newblock Topicrank: Graph-based topic ranking for keyphrase extraction.
\newblock In \emph{International joint conference on natural language processing (IJCNLP)}, pages 543--551, 2013.

\bibitem[Duari and Bhatnagar(2019)]{duari2019scake}
Swagata Duari and Vasudha Bhatnagar.
\newblock scake: semantic connectivity aware keyword extraction.
\newblock \emph{Information Sciences}, 477:\penalty0 100--117, 2019.

\bibitem[Grootendorst(2021)]{zenodo}
Maarten Grootendorst.
\newblock Maartengr/keybert.
\newblock \url{https://zenodo.org/record/4461265}(Accessed: 3 October 2023), 2021.
\newblock URL \url{https://zenodo.org/record/4461265}.

\end{thebibliography}

\end{document}